\pdfoutput=1
\documentclass{bmvc2k}
\usepackage{enumitem}
\usepackage{multirow}
\usepackage{amssymb}
\usepackage{bm}

\newcommand{\ak}{\textcolor{black}}

\title{Story Understanding in Video Advertisements}

\addauthor{Keren Ye}{yekeren@cs.pitt.edu}{1}
\addauthor{Kyle Buettner}{buettnerk@pitt.edu }{1}
\addauthor{Adriana Kovashka}{kovashka@cs.pitt.edu}{1}

\addinstitution{
 Department of Computer Science\\
 University of Pittsburgh\\
 Pennsylvania, United States
}

\runninghead{\scriptsize{Keren Ye, Kyle Buettner, Adriana Kovashka}}{\scriptsize{Story Understanding in Video Advertisements}}

\begin{document}

\maketitle

\begin{abstract}
In order to resonate with the viewers, many video advertisements explore creative narrative techniques such as ``Freytag's pyramid'' where a story begins with exposition, followed by rising action, then \emph{climax}, concluding with denouement. In the dramatic structure of ads in particular, climax depends on changes in sentiment.
We dedicate our study to understand the dynamic structure of video ads automatically. To achieve this, we first crowdsource climax annotations on 1,149 videos from the Video Ads Dataset, which already provides sentiment annotations. We then use both unsupervised and supervised methods to predict the climax. Based on the predicted peak, the low-level visual and audio cues, and semantically meaningful context features, we build a sentiment prediction model that outperforms the current state-of-the-art model of sentiment prediction in video ads by 25\%.
In our ablation study, we show that using our context features, and modeling dynamics with an LSTM, are both crucial factors for improved performance.
\end{abstract}

\vspace{-0.5cm}
\section{Introduction}
\label{sec:intro}
\vspace{-0.2cm}

Video advertisements are powerful tools for affecting the public opinion, by appealing to the viewers' emotions \cite{young2008advertising}.
To achieve persuasive power, many ads explore creative narrative techniques. One classic technique is ``Freytag's pyramid'' where a story begins with exposition (setup), followed by rising action, then climax (action and sentiment peak), concluding with denouement or resolution (declining action) \cite{freytag1896freytag}.

\begin{figure}
    \centering
    \includegraphics[width=1\linewidth]{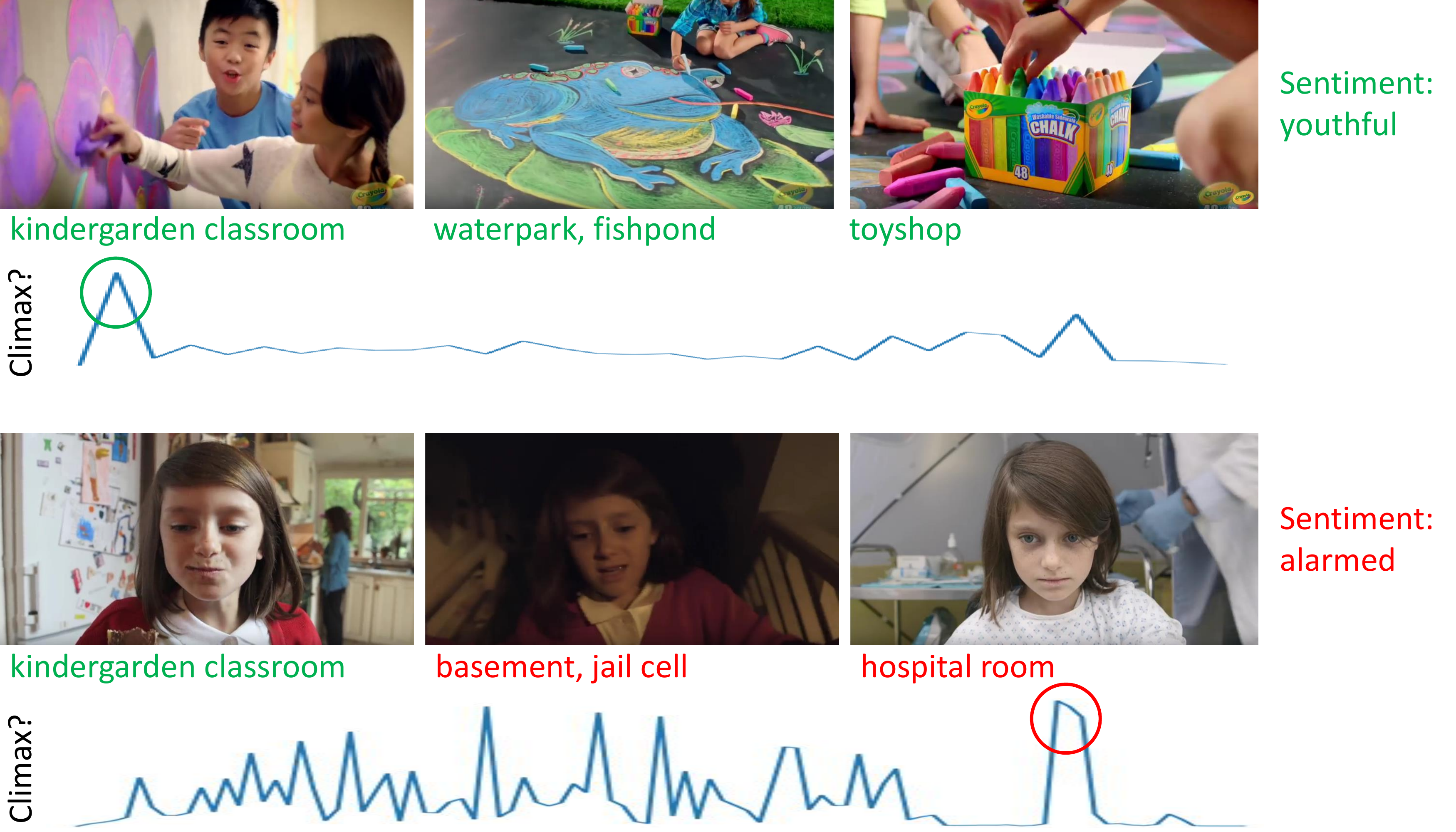}
    \vspace{-0.5cm}
    \caption{The key idea behind our approach. We want to understand the story being told in the ad video, and the sentiment it provokes. We hypothesize that the semantic content of each frame is quite informative and that we need to model the rising action to understand which temporal parts most contribute to the sentiment. We show the places recognized in the frames of two videos, as well as soft predictions about whether a certain frame corresponds to the climax of the video or not. While both videos start with images of children, which might indicate positive sentiment denoted in green (e.g. ``youthful''), this positive trend only remains in the first video (indicated by places correlated with youthfulness, such as ``toy shop''). In contrast, the second video changes course and shows unpleasant places (denoted in red) e.g. ``basement'' and ``hospital room''. Because the climax in the second video occurs near the end, our method understands that it is these later frames that determine the sentiment (``alarmed'').} 
    \label{fig:concept}
\end{figure}

In this work, we model the dynamic structure of a video ad. We track the pacing and intensity of the video, using both the visual and audio domains. We model how emotions change over the course of the ad. We also model correlations between specific settings (e.g., child's bedroom), objects (e.g., teddy bear) and sentiments (e.g., happy). 
We propose two methods to predict \emph{climax}, ``the highest dramatic tension or a major turning point in the action'' \cite{mw}, of a video. Then we use them along with rich context features to predict the \emph{sentiment} that the video provokes in the viewer. 
Our framework is illustrated in Fig.~\ref{fig:concept}.
Our techniques are based on the following two hypotheses which we verify in our experiments. 

First, we hypothesize that the climax of a video correlates with dramatic visual changes or intense content. Thus, we compute optical flow per frame and detect shot boundaries, then predict that climax occurs at those moments in the video where peaks in optical flow vectors or shot boundary changes occur. To measure dynamics in the audio domain, we extract the amplitude of the sound channel and predict climax when we encounter peaks in the amplitude.
In addition to this unsupervised approach, we also show how to use the cues we develop as features, to predict climax in a supervised way.
Both the unsupervised and supervised approaches greatly outperform the baseline tested.

Second, we hypothesize that video ads exploit associations that humans make, to create an emotional effect. We aim to predict the sentiment that an ad provokes in the viewer, and hypothesize that the setting and objects in the ad are greatly responsible for the sentiment evoked. We first extract predictions about the type of scene and type of objects in the ad, for each frame.
We also hypothesize that the facial expressions of the subjects of the ad (i.e., the people in the ad) correlate with the sentiment provoked in the people watching it, so we also extract per-frame facial expression predictions. 
We treat sentiment prediction as a recurrent prediction task based on the scene, object, and emotion features, as well as features related to climax and standard ResNet \cite{he2016deep} visual features. 


To train our methods and test our hypotheses, we crowdsource climax annotations on 1,149 videos from the Video Ads Dataset of \cite{hussain2017automatic}, and use the sentiment annotations provided.

\section{Related Work}
\label{sec:related}
\vspace{-0.3cm}

\emph{Video dynamics and actions.}
Optical flow  \cite{fleet2006optical,brox2004high,sun2010secrets,mayer2016large,ilg2017flownet,ranjan2017optical} is a basic building block of video understanding. We use \cite{ranjan2017optical} due to its simplicity and reliable accuracy. 
Higher-level analysis of video includes human pose estimation \cite{toshev2014deeppose,shotton2013efficient,newell2016stacked} and action detection and recognition \cite{yeung2016end,gkioxari2015finding,wang2015action,carreira2017quo}. 
Unlike these, optical flow does not capture semantics (such as the name of the action performed in a video). This is desirable in our case since a wide variety of activities can be exciting and climactic, so categorization is less useful. Anomaly detection \cite{mahadevan2010anomaly} is also related, but rather than predicting what does not fit, we wish to predict how a video builds up and increases its dramatic content to create the climax.

\emph{Emotions.}
Researchers have been interested in predicting facial expressions and emotions for a long time \cite{essa1997coding,kanade2000comprehensive,cohen2003facial}. Large datasets exist \cite{mollahosseini2017affectnet,benitez2016emotionet,kosti2017emotic}. We train a facial expression model on \cite{mollahosseini2017affectnet} and apply it on faces detected in the video, as a cue for the viewers' sentiment. 

\emph{Movie and story understanding.}
We attempt to understand the stories told by video ads. Others have previously developed techniques for understanding various aspects of movies, such as their plot \cite{tapaswi2016movieqa, Na_2017_ICCV} and the principal characters and their relations \cite{weng2009rolenet}. 
While there is no prior work on detecting climax in ads, some previous approaches model the tempo of other videos.
For example, \cite{liu2008innovative} use cues like ``motion intensity'' and ``audio pace'' to detect action scenes. 
\cite{rasheed2002movie} use the pacing of a movie to recognize its genre (action movies are faster-paced than dramas).
\cite{choi2016video} create video stories out of consumer videos, using story composition, dynamics and coherency, as cues.
However, these works do not take emotions nor context such as scene and surrounding objects into account. We show semantic context features improve the performance of the unsupervised cues (e.g. ``motion intensity'').

\emph{Advertisement and media understanding.}
There is a recent trend to attempt to understand the visual media with computer vision techniques. 
\cite{joo2014visual,hussain2017automatic,ye2018advise,won2017protest} analyze the hidden messages of images, in news articles \cite{joo2014visual,won2017protest} and advertisements \cite{hussain2017automatic}. \cite{joo2015automated,wang2017polarized} examine the visual distinctions between people either running or voting in elections. 
We use the dataset of \cite{hussain2017automatic} for our study, and show that we greatly outperform their sentiment prediction model.

\vspace{-0.5cm}
\section{Approach}
\label{sec:approach}
\vspace{-0.2cm}

\begin{figure}
    \centering
    \includegraphics[width=1\linewidth]{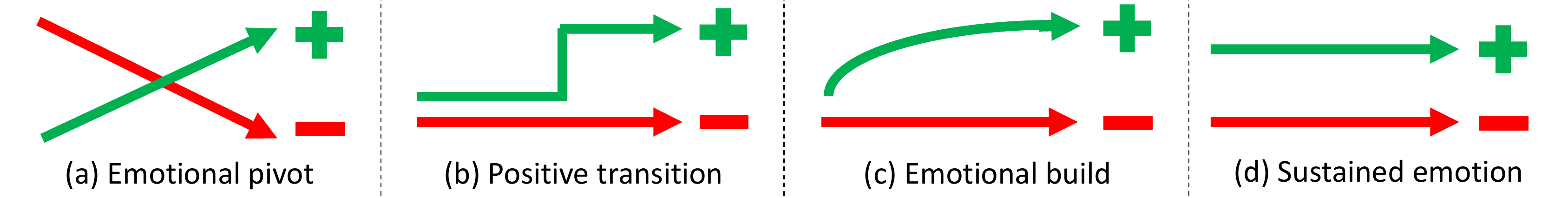}
    \caption{The ``four archetypes of dramatic structure'' in product ads \cite{young2008advertising} which motivate our approach. For PSAs, the roles of positive and negative sentiments might be reversed.}
    \label{fig:archetypes}
\end{figure}

In The Advertising Research Handbook \cite{young2008advertising}, dramatic structure has four prototypical forms, shown in Fig.~\ref{fig:archetypes} (based on \cite{young2008advertising} p.212).
These structures depend on how positive and negative sentiment rises or declines. \cite{young2008advertising} examines product ads, and the changes in positive/negative sentiment are correlated with appearances of the brand. In public service announcements (PSAs), the role of positive/negative might be reversed, as PSAs often aim to create negative sentiment in order to change a viewer's behavior. However, understanding the story of PSAs still depends on understanding the climax of (negative) sentiment. 
Thus, we first collect data (Sec.~\ref{sec:data}) and develop features (Sec.~\ref{sec:climax}) that help us predict when climax occurs. We then develop features informative for sentiment (Sec.~\ref{sec:sent}). We finally describe how we use these features to predict the type of sentiment and  occurrences of climax (Secs.~\ref{sec:peaks} and \ref{sec:sup}).

\vspace{-0.5cm}
\subsection{Climax and sentiment data}
\label{sec:data}
\vspace{-0.2cm}

We use the Video Ads Dataset of \cite{hussain2017automatic}. It contains 3,477 video advertisements with a variety of annotations, including the sentiment that the ad aims to provoke in the viewer. We collected climax annotations on a randomly chosen subset of 1,595 videos from this dataset, using the Amazon Mechanical Turk platform. 
We restricted participation on our tasks to annotators with at least 98\% approval rate who submitted at least 1000 approved tasks in the past. 
We submitted each video for annotation to four workers. Each was asked to watch the video and could choose between two options, ``the video has no climax'' or ``the video has climax.'' If the latter, the worker was asked to provide the minute and second at which climax occurs (most videos are less than 1 min long). To ensure quality, annotators were also asked to describe what happens at the end of the video. Some of the videos in \cite{hussain2017automatic}'s dataset were not available, so the annotators could also mark this option. We ended up with 1,149 videos that contain climax annotations.
We manually inspected a subset of them and found the timestamps were quite reasonable. The descriptions of what happened at the end were often quite detailed.  
We will make this data publicly available upon publication.

\begin{figure}[t]
    \centering
    \includegraphics[width=1\linewidth]{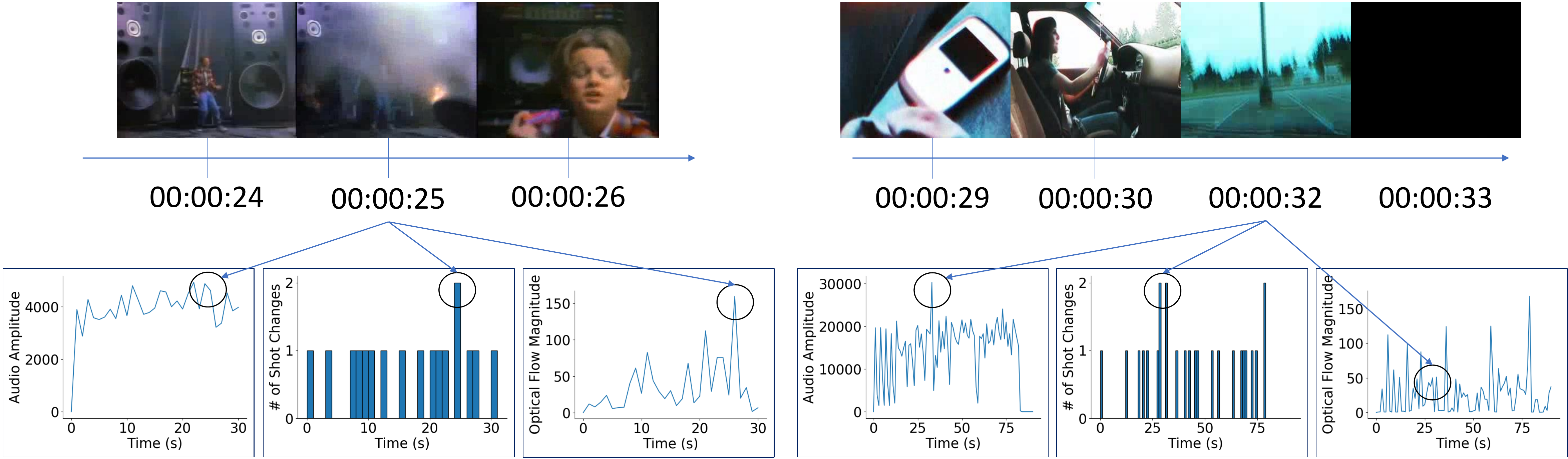}
    \caption{The audio, shot boundary \ak{frequency}, and optical flow plots for two videos, along with frames from the videos corresponding to climactic points. The first video shows an ``explosion'' around the 25th second, and the second shows a car crash around the 32nd second. The circles correspond to the timestamp of the frames shown. In the first video, climax is detected well in each of the three plots. In the second, shot boundaries and audio are informative, but optical flow is not.}
    \label{fig:plots}
\end{figure}

\vspace{-0.5cm}
\subsection{Climax indicators}
\label{sec:climax}
\vspace{-0.2cm}

We first analyze the dynamics of the video, using both visual and audio channels. We plot time on the x-axis, and measurement of dynamics/activity on the y-axis (Fig.~\ref{fig:plots}). We consider three indicators of rapid activity: the amplitude of audio signals, the occurrence of shot boundaries, and the magnitude of optical flow vectors between frames.

In particular, we extract these features and portray them as follows: 
\begin{itemize}[noitemsep,nolistsep]
    \item \textbf{The audio amplitude $\bm{a}^k$}, which is the max amplitude of audio for the $k$-th frame. We first extract the sound channel from the video, take a fixed number of samples from the sound wave per second, then compute the max across the samples for that frame. 
    \item \textbf{The shot boundary indicator}, which is equal to 0 or 1 depending on whether a shot boundary occurs in the $k$-th frame. We use \cite{sbd} for shot boundary extraction. In order to obtain more informative cues, we vary the parameters of \cite{sbd} to get five 0/1 predictions per frame and use this 5D prediction $\bm{b}^k$ as the representation for the $k$-th frame. To generate the plot in Fig.~\ref{fig:plots}, we aggregate information over all frames in a given second.
    \item \textbf{The optical flow magnitude $\bm{o}^k$}, which is computed as $\frac{1}{W*H} \sum_{i=1}^W \sum_{j=1}^H \sqrt{u^{k \; 2}_{i,j} + v^{k \; 2}_{i,j}}$ where $u^k_{i,j}$ and $v^k_{i,j}$ are the horizontal and vertical optical flow components for each pixel $(i, j)$ in the $k$-th frame. We use \cite{ranjan2017optical} to extract optical flow vectors.

\end{itemize}


\vspace{-0.5cm}
\subsection{Sentiment indicators}
\label{sec:sent}
\vspace{-0.2cm}

The Advertising Research Handbook \cite{young2008advertising} describes the dramatic structure of ads as closely depending on the emotion of the video. 
One type of structure (Fig.~\ref{fig:archetypes}) is the ``emotional pivot'' where an ad starts with negative sentiment, which declines over time, to make room for increasing positive sentiment. The ``emotional build'' involves a gradual increase and climax in positive sentiment. Thus, the sentiment is equally crucial to understanding the story of the ad video as the climax. 
Since an ad targets an audience and wants to convince the audience to do something, it is the viewer's sentiment that matters the most. 

\begin{figure}[t]
    \centering
    \includegraphics[width=1\linewidth]{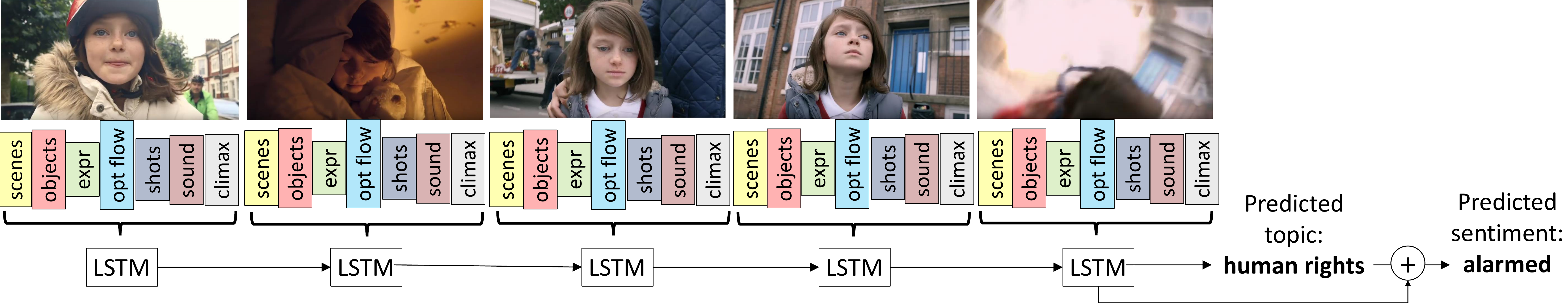}
\vspace{-0.7cm}
    \caption{Our dynamic context-based approach. The last frame shows an explosion.}
\vspace{-0.5cm}
    \label{fig:approach}
\end{figure}

\cite{hussain2017automatic} contains annotations about what sentiment each ad video provokes in the viewer, collected from five annotators. These annotations involve 30 sentiments, both positive (e.g., cheerful, inspired, educated), negative (e.g., alarmed, angry) and neutral (e.g., empathetic).
\cite{hussain2017automatic} also includes a baseline for predicting sentiment, using a multi-class SVM and C3D features \cite{tran2015learning}.
The authors extract features from 16-frame video clips, then average the features. 
Thus, their model does not capture the dynamics and sequential nature of the video.
We hypothesize that if we model how the content of the video changes \emph{over time}, and consider the \emph{context} in which the sentiment in the video is conveyed, we would be able to model sentiment more accurately.
We model sentiment with the following intuitive context features:
\begin{itemize}[noitemsep,nolistsep]
    \item \textbf{The setting in each frame of the video}, i.e. the type of place/scene. Let $\textbf{\emph{vp}} = \{p_1, \dots, p_{365}\}$ be the vocabulary of places in the Places365 dataset \cite{zhou2017places}. We use a pre-trained prediction model from \cite{zhou2017places} to obtain a 365D vector $\bm{pl}^k=[l_1^k, \dots, l_{365}^k]$, where $l_i^k$ is the probability that the $k$-th frame exemplifies the $i$-th place.
    
    \item \textbf{The objects found in the video}. 
    Let $\textbf{\emph{vo}}=\{c_1, c_2, \dots,c_{80}\}$ be the vocabulary of the COCO object detection dataset \cite{lin2014microsoft}. We use the model of \cite{Huang_2017_CVPR} trained on COCO to get the objects in a frame. We then use max-pooling to turn the detection results into an 80D fixed-length feature vector $\bm{ob}^k=[s_1^k, \dots, s_{80}^k]$, where $s_i^k$ is the maximum confidence score among multiple instances of the same object class $c_i$, in frame $k$.
    
    \item \textbf{The facial expressions in the video}. We observed that the overall sentiment that the video provokes \emph{in the viewer} often depends on the emotions that the \emph{subjects} of the video go through. For example, if a child in an ad video is initially ``happy'' but later becomes ``sad,'' the sentiment provoked in the adult viewer might be ``alarmed'' because something disturbing must have happened. Thus, we also model emotions predicted on faces extracted per frame. 
    We first detect the faces using  OpenFace \cite{amos2016openface}. We then extract the expression of each face using an Inception model \cite{szegedy2016rethinking} trained on the AffectNet dataset \cite{mollahosseini2017affectnet}. Two types of results are predicted: (1) the probability distribution among the eight expressions defined in AffectNet, and (2) the valence-arousal values for the face, saying how pleased and how active the person is (in range -1 to +1). We average the face expressions (10 values) for all faces detected in the $k$-th frame, to get the 10D final representation $\bm{fa}^k$.
    
    \item \textbf{The topic of the ads}. \cite{hussain2017automatic} defines a vocabulary of 38 topics in the ads domain and also provides annotations for these topics. We hypothesize the overall sentiment that the video provokes is related to the topic the ad belongs to. For example, ``sports'' ads usually convey ``active'' and ``manly'' sentiments, while ``domestic violence'' ads often make people feel ``sad''. \ak{Thus, we designed a multi-task learning framework with two objectives: one for the topic and the other for the sentiment prediction, hoping the topic prediction can help the prediction of sentiments. We first use the video-level feature (the last hidden state of the LSTM) to predict the 38D topic distribution, then concatenate this 38D vector with the video-level feature to predict the sentiment. The idea is described in Fig.~\ref{fig:approach}.}
    
\end{itemize}

We also use features from the last layer of a ResNet trained on ImageNet \cite{he2016deep,russakovsky2015imagenet}. 

\subsection{Unsupervised climax prediction}
\label{sec:peaks}
\vspace{-0.2cm}

We can directly predict that climax occurs at times which are peaks in terms of shot boundary frequency, optical flow magnitude, or audio amplitude. 
Since the shot boundary frequency can be the same for many timeslots, we look for the longest sequence of timeslots which contain at least one shot boundary and predict the center of this ``run'' as a peak. Optical flow magnitudes and audio amplitudes are compared on a second-by-second basis.
We extract the top-$k$ maximal responses from each plot, predict these as climax, and evaluate the performance in Sec.~\ref{sec:climax_eval}.

\vspace{-0.5cm}
\subsection{Supervised prediction}
\label{sec:sup}
\vspace{-0.2cm}

We predict climax using an LSTM (with 64 hidden units) that outputs 0/1 for each frame, where 1 denotes that the frame is predicted to contain climax. The frame-level features used are ResNet features (2048D), optical flow magnitude $\bm{o}^k$ (1D), the shot boundary indicator $\bm{b}^k$ (5D), the sound amplitude $\bm{a}^k$ (1D), the place representation $\bm{pl}^k$ (365D), the object representation $\bm{ob}^k$ (80D), and the facial expression feature $\bm{fa}^k$ (10D).

For the sentiment prediction task, we also use an LSTM with 64 hidden units. We use the same frame-level features as the climax prediction. Moreover, we also add the predicted climax (1D) as extra information. Ads topics are used as both an additional loss/constraint and an extra feature for the sentiment prediction (see Fig.~\ref{fig:approach}).



\vspace{-0.5cm}
\subsection{Discussion}
\vspace{-0.2cm}


The advantages of our approach are as follows. First, the distribution of object, place, and facial expression probability vectors is much lower-dimensional than ResNet features, so given the limited size of the Video Ads Dataset (3,477), formulating the problem as learning a mapping from objects/scenes/facial expressions to sentiments/climax is much more feasible. 
The optical flow, shot boundary, and sound features are also very low-dimensional, and have clear correlation with the presence of climax. Further, understanding the sentiment of a video and its climax are related tasks. Thus, it is intuitive that climax predictions should be allowed to affect sentiment prediction; this is the idea shown in Fig.~\ref{fig:concept} where we use climax to select the part of the video which affects the elicited sentiment the most.
We show in Sec.~\ref{sec:sent_eval} (Table \ref{tab:ablations}) that our semantic/climax features outperform the ResNet features, and the combination of the two achieves the strongest performance. 




\vspace{-0.5cm}
\section{Experimental Validation}
\label{sec:results}
\vspace{-0.2cm}

We first describe our experimental setup and training procedure, then present quantitative and qualitative results on the climax and sentiment prediction tasks.

\vspace{-0.5cm}
\subsection{Evaluation metrics}
\vspace{-0.2cm}

For the climax prediction task, we use the recall of the top-$k$ prediction ($k=1,3$) to measure performance. Since exactly matching the ground-truth climax timestamp is challenging, we apply an error window saying that the prediction is treated as correct if the ground-truth climax is close (within $0,1,2$ sec). \ak{We treat the prediction as correct if it recalls any of the ground-truth annotations} for that video, except rejected work. Table \ref{tab:climax} shows the results.

To measure how well the model's prediction agrees with the sentiment annotations, we compute mean average precision (mAP) and top-1 accuracy (acc@1) based on three forms of agreement (agree with $k$, where $k=1,2,3$). ``Agree with $k$'' means that we assign a ground-truth label to a video only if at least $k$ annotators agree on the existence of the sentiment. The acc@1 is the fraction of correct top-1 predictions across all videos, and the mAP is the mean of the average precision over evenly spaced recall levels. Tables \ref{tab:main}, \ref{tab:secondary} and \ref{tab:ablations} show the results.


\vspace{-0.5cm}
\subsection{Training and implementation details}
\vspace{-0.2cm}

For training both the climax and sentiment prediction models, we use the TensorFlow \cite{abadi2016tensorflow} deep learning framework. We split the Video Ads Dataset \cite{hussain2017automatic} (3,477 videos) into train/val/test (60\%/20\%/20\%), resulting in around 2,000 training examples for the sentiment prediction task and about 700 training examples for the climax prediction task (since only 1,149 of the 3,477 videos have climax annotation). We report our results using five-fold cross-validation.

For the climax prediction task, we use a one-layer LSTM model with 64 hidden units. \ak{At each timestamp, the model predicts a real value ranging from $[0,1]$ (output of the sigmoid function) denoting whether the corresponding frame contains a climax. We then use the sigmoid cross entropy loss to constrain the model to mimic the human annotations.} Considering the size of the dataset, we set both the input and output dropout keep probability of the LSTM cell to 0.5 to avoid over-fitting. We use the RMSprop optimizer with a decay factor of 0.95, momentum of 1e-8, and learning rate of 0.0002. We train for 20,000 steps using a batch size of 32, and we use the recall of the top-1 prediction (the error window is set to ``within 2 seconds'') to pick the best model on the validation set.

For the sentiment prediction task, we use the same procedure, but we pick the best model using mAP using ``agreement with 2''.
We use the last hidden state of the LSTM to represent the video feature and add a fully connected layer upon it to get the 38D topic representation. We then concatenate the 38D topic representation with the last hidden state of the LSTM and infer a 30D sentiment logits vector from the concatenated feature. The sigmoid cross entropy loss is also used here. Similar to \cite{teney2017tips}, we found that using soft scores as ground-truth targets improves the performance and makes the training more stable. To deal with data imbalance for the rare classes, we sampled at most $5n$ negative samples if there were $n$ positives.  

\vspace{-0.4cm}
\subsection{Climax prediction}
\label{sec:climax_eval}
\vspace{-0.2cm}

We show the results of unsupervised and supervised climax prediction in Table \ref{tab:climax}. We measure whether the predicted climax is within 0, 1, or 2 seconds of the ground-truth climax. We first show a heuristic-guess baseline which always predicts that climax occurs at 5 seconds for the top-1 prediction and at 5, 15 and 25 seconds for top-3. We then show the performance of the three unsupervised climax prediction methods described in Sec.~\ref{sec:peaks}. Next, we show the performance of 0/1 climax prediction (Sec.~\ref{sec:sup}) using an LSTM with ResNet features only, and finally our method using the features we proposed in both Sec.~\ref{sec:climax} and Sec.~\ref{sec:sent} (excluding the video-level topic feature). 

\begin{table}[t]
    \centering
    \begin{tabular}{|c||c|c|c||c|c|c|}
\hline
&	\multicolumn{3}{c||}{top-1 prediction} &		\multicolumn{3}{c|}{top-3 prediction} \\
\hline
Method & w/in 0 s & w/in 1 s & w/in 2 s & 	w/in 0 s & w/in 1 s & w/in 2 s\\
\hline
\hline
baseline & 0.031 & 0.083 & 0.121 & 0.122 & 0.299 &	0.430\\
\hline
\hline
shot boundary (unsup) &	0.068 &	0.179 &	0.265 &	\emph{0.221} &	\textbf{0.457} &	\textbf{0.588}\\
\hline
optical flow (unsup) &	0.064 &	0.152 &	0.220 &	0.163 &	0.380 &	0.513\\
\hline
audio (unsup) & \textbf{0.077} &	0.171 &	0.255 &	0.178 &	0.403 &	0.534\\
\hline
\hline
LSTM, ResNet only &	0.071 &	\emph{0.206} &	\textbf{0.290} &	0.190 &	0.400 &	0.523\\
\hline
LSTM, all feats (Ours) &	\textbf{0.077} &	\textbf{0.209} &	\emph{0.287} &	\textbf{0.226} &	\emph{0.439} &	\emph{0.546}\\
\hline
    \end{tabular}
    \caption{Climax prediction with best performer per setting in \textbf{bold} and second performer in \emph{italics}. Unsupervised prediction performs quite well. Our supervised method achieves the best or second-best performance for all settings. \ak{For the ``LSTM, ResNet only'' approach, we guess the reason that it is competitive is that LSTM has the ability to capture the temporal dynamics to a certain degree.}} 
    \label{tab:climax}
\end{table}

We see that the unsupervised methods, and especially shot boundary and audio, greatly outperform the baseline. Interestingly, audio performs quite well in the hardest setting, only one shot at prediction and exact alignment between predicted and ground-truth climax. Shot boundary achieves the best performance in the two weakest settings (top-3 predictions, agreement within 1-2 seconds). In all settings, our method achieves the best or second-best performance.

\vspace{-0.4cm}
\subsection{Sentiment prediction}
\label{sec:sent_eval}
\vspace{-0.2cm}

Table \ref{tab:main} shows our main result for sentiment prediction. We compare to Hussain et al. \cite{hussain2017automatic}'s method which is a multi-class SVM model using the C3D features \cite{tran2015learning}. This is the only prior method that attempts to predict sentiment on the Video Ads Dataset.
We observe that our method improves upon \cite{hussain2017automatic}'s performance for most metrics. The improvement is more significant for mAP, which is more reliable because of the imbalance of the dataset. We improve the mAP compared to prior art by up to 25\% in terms of agreement with 3 annotators.
For reference, human annotators' agreement with 1 (at least one other annotator) is 0.723.

\begin{table}[t]
    \centering
    \begin{tabular}{|c||c|c||c|c||c|c|}
    \hline
     & \multicolumn{2}{c||}{Agree with 1} & \multicolumn{2}{|c||}{Agree with 2} & \multicolumn{2}{|c|}{Agree with 3}\\
    \hline
    Method & mAP & acc@1 & mAP & acc@1 & mAP & acc@1 \\
    \hline
    \hline
    Hussain et al. \cite{hussain2017automatic} & 0.283	& 0.664	& 0.135	& 0.435	& 0.075 & \textbf{0.243}\\
    \hline
    Our model & \textbf{0.313} & \textbf{0.712} & \textbf{0.160} & \textbf{0.449} & \textbf{0.094} & 0.241 \\
    \hline
    \end{tabular}
    \caption{Our method outperforms prior art for sentiment prediction.}
    \label{tab:main}
    \vspace{-0.3cm}
\end{table}

Table \ref{tab:secondary} examines the contribution of the features described in Sec.~\ref{sec:climax} and Sec.~\ref{sec:sent}, and the use of an LSTM to model dynamics of the video.
We compare against an LSTM that uses only ResNet features.
We also compare to a bag-of-frames (BOF) method that rules out the effects of dynamics. It computes the final video-level representation by simply applying mean pooling among the frame-level features.
We observe that our method (using the proposed features and LSTM) always outperforms the other methods in terms of mAP scores. Our method achieves significant improvement over the second-best method (10\% for mAP and agreement with 2, and 21\% for mAP and agreement with 3). In terms of accuracy, all methods perform similarly, and the best model (BOF, all features) also uses our proposed features. 

\begin{table}[t]
    \centering
    \begin{tabular}{|c||c|c||c|c||c|c|}
    \hline
     & \multicolumn{2}{c||}{Agree with 1} & \multicolumn{2}{|c||}{Agree with 2} & \multicolumn{2}{|c|}{Agree with 3}\\
    \hline
    Method & mAP & acc@1 & mAP & acc@1 & mAP & acc@1 \\
    \hline
   BOF, ResNet only & 0.295 & 0.708 & 0.141 & 0.449 & 0.076 & 0.242 \\
    \hline
    LSTM, ResNet only & 0.302 & 0.716 & 0.145 & 0.451 & 0.074 & 0.242 \\
    \hline    
    BOF, all features (incl. ours) & 0.302 & \textbf{0.719} & 0.146 & \textbf{0.462} & 0.078 & \textbf{0.248} \\
    \hline
    LSTM, all features (Our model) & \textbf{0.313} & 0.712 & \textbf{0.160} & 0.449 & \textbf{0.094} & 0.241 \\
    \hline
    \end{tabular}
    \caption{In-depth evaluation of the components of our method for sentiment prediction.}
    \label{tab:secondary}
\end{table}

Table \ref{tab:ablations} verifies the benefit of each of our features. We show the LSTM-ResNet-only baseline from Table \ref{tab:secondary}, then eight methods which add one of our features at a time, on top of this baseline. Next, we show an LSTM method which uses our features without the base ResNet feature, and finally, our full method.
We use mAP for agreement with 3 in the table. We show the average result across all sentiment classes, then results for four individual ad sentiments. In bold are all methods which improve upon the ResNet baseline. We see that all of our features (the average column) contribute to the performance of our full method. 
Using all features except ResNet is stronger than using ResNet features alone. \ak{We note models based on individual features still show benefits on specific sentiment classes, and we believe the reason is that our fusion method is too simple to aggregate all the information.}

\begin{table}[t]
    \centering
    \begin{tabular}{|c||c||c|c|c|c|}
\hline
& \textbf{average} &	educated &	alarmed &	fashionable	& angry\\
\hline
\hline
ResNet only (baseline) &	0.074 & 0.036 & 0.117 & 0.047 & 0.007\\
\hline
\hline
objects &	\textbf{0.082} & 0.032 & \textbf{0.140} & \textbf{0.080} & 0.004 \\
\hline
places &	\textbf{0.082} & \textbf{0.074} & \textbf{0.132} & \textbf{0.160} & 0.005 \\
\hline
facial expressions &	\textbf{0.077} & \textbf{0.044} & \textbf{0.143} & \textbf{0.084} & 0.003\\
\hline
topic &	\textbf{0.086} & 0.032 & \textbf{0.143} & \textbf{0.136} & \textbf{0.009}\\
\hline
optical flow &	\textbf{0.082} & \textbf{0.045} & \textbf{0.150} & \textbf{0.133} & 0.005 \\
\hline
shot boundaries &	\textbf{0.080} & \textbf{0.037} & \textbf{0.151} & \textbf{0.110} & 0.003\\
\hline
audio &	\textbf{0.077} & \textbf{0.040} & 0.113 & \textbf{0.116} & \textbf{0.010} \\
\hline
climax &	\textbf{0.079} & 0.025 & \textbf{0.119} & \textbf{0.082} & \textbf{0.011} \\
\hline
\hline
all features except ResNet &	\textbf{0.080} & \textbf{0.038} & 0.104 & 0.036 & 0.007 \\
\hline
\hline
all features (Our model) &	\textbf{0.094} & 0.026 & 0.099 & \textbf{0.202} & 0.005 \\
\hline
    \end{tabular}
    \caption {Ablation study evaluating the benefit of each feature for sentiment prediction, using agreement with 3 mAP. In bold are all methods that outperform the baseline. }
    \label{tab:ablations}
    \vspace{-0.5cm}
\end{table}

We observe some intuitive results for the four chosen individual sentiments. We ranked sentiments by frequency in the dataset and picked the 6th, 7th, 9th and 13th most frequent. For ``educated,'' the places feature is most beneficial, which makes sense because ``education'' might occur in particular environments, e.g., classroom. 
As shown in our example ad in Fig.~\ref{fig:approach}, the setting (e.g., places) and dramatic content changes (measured by optical flow and shot boundaries) are quite telling of the ``alarmed'' sentiment. 
Most features help greatly for the ``fashionable'' sentiment. 
For ``angry'', audio is very helpful (43\% improvement over ResNet), which makes sense since loud speaking might trigger or correlate with anger. 

\begin{figure}[t]
    \centering
    \includegraphics[width=0.95\linewidth]{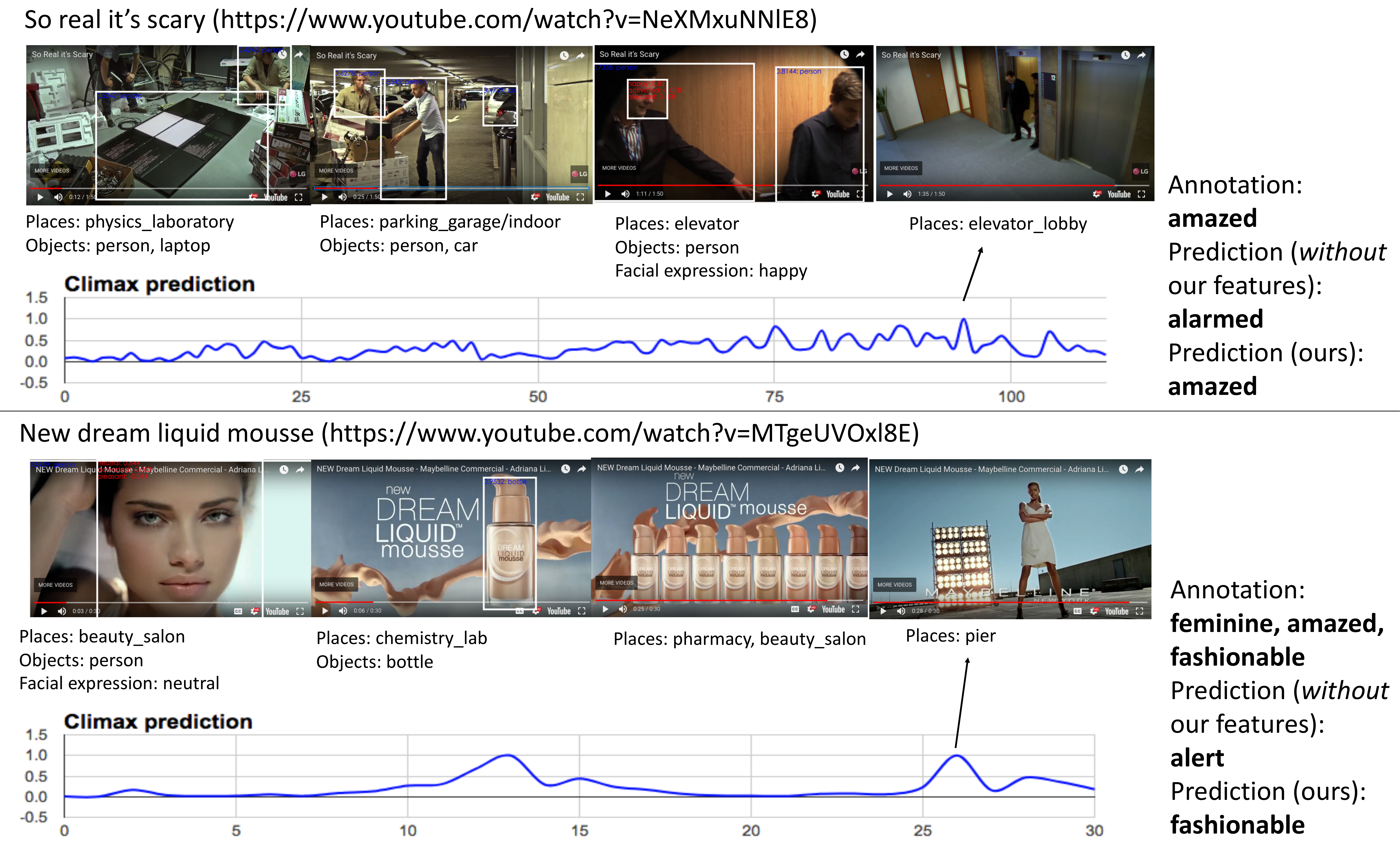}
    \caption{Qualitative results from our model. }
    \label{fig:quali}
\end{figure}

We show qualitative examples in Fig.~\ref{fig:quali}. 
Our model's features correctly predict ``amazed'' and ``fashionable'' while the baseline method does not. Our method relies on recognized places (e.g. laboratory, beauty salon), objects, facial expressions, and climax dynamics.

\section{Conclusion}
\label{sec:conclusion}
\vspace{-0.5cm}

We made encouraging progress in understanding the dynamic structure of a video ad. We hypothesized that climax correlates with dramatic visual and audio changes. We crowdsourced climax annotations on 1,149 videos from the Video Ads Dataset of \cite{hussain2017automatic} and used both unsupervised and supervised methods to predict the climax. 
By combining visual and audio cues with semantically meaningful context features, our sequential model (LSTM) outperforms the only prior work \cite{hussain2017automatic}   by a large margin. To better understand the relations between the semantic visual cues and the sentiment each ad video provokes, we performed detailed ablations and found all the features we proposed help to understand the evoked sentiment. 
In the future, we will investigate other resources relevant to both climax and sentiment in video ads.
We will also improve the interpretability of the model.
\ak{Finally, our ablation studies show the limitation of the feature fusion method of our model thus we will investigate additional fusion strategies to further improve the performance.}
\section*{Acknowledgments}
\label{sec:ack}
\vspace{-0.3cm}

This material is based upon work supported by the National Science Foundation under Grant Number 1566270. This research was also supported by a Google Faculty Research Award and an NVIDIA hardware grant. Any opinions, findings, and conclusions or recommendations expressed in this material are those of the author(s) and do not necessarily reflect the views of the National Science Foundation. The authors also appreciate the help of Sanchayan Sarkar and Chris Thomas for preparing the data for and training the facial expression models.


\bibliography{egbib}

\begin{thebibliography}{46}
\providecommand{\natexlab}[1]{#1}
\providecommand{\url}[1]{\texttt{#1}}
\expandafter\ifx\csname urlstyle\endcsname\relax
  \providecommand{\doi}[1]{doi: #1}\else
  \providecommand{\doi}{doi: \begingroup \urlstyle{rm}\Url}\fi

\bibitem[mw()]{mw}
Merriam-webster.com.
\newblock \url{https://www.merriam-webster.com/dictionary/climax}.

\bibitem[Abadi et~al.(2016)Abadi, Barham, Chen, Chen, Davis, Dean, Devin,
  Ghemawat, Irving, Isard, et~al.]{abadi2016tensorflow}
Mart{\'\i}n Abadi, Paul Barham, Jianmin Chen, Zhifeng Chen, Andy Davis, Jeffrey
  Dean, Matthieu Devin, Sanjay Ghemawat, Geoffrey Irving, Michael Isard, et~al.
\newblock Tensorflow: A system for large-scale machine learning.
\newblock In \emph{OSDI}, volume~16, pages 265--283, 2016.

\bibitem[Amos et~al.(2016)Amos, Ludwiczuk, and
  Satyanarayanan]{amos2016openface}
Brandon Amos, Bartosz Ludwiczuk, and Mahadev Satyanarayanan.
\newblock Openface: A general-purpose face recognition library with mobile
  applications.
\newblock \emph{CMU School of Computer Science}, 2016.

\bibitem[Benitez-Quiroz et~al.(2016)Benitez-Quiroz, Srinivasan, Martinez,
  et~al.]{benitez2016emotionet}
Carlos~Fabian Benitez-Quiroz, Ramprakash Srinivasan, Aleix~M Martinez, et~al.
\newblock Emotionet: An accurate, real-time algorithm for the automatic
  annotation of a million facial expressions in the wild.
\newblock In \emph{Proceedings of the IEEE Conference on Computer Vision and
  Pattern Recognition (CVPR)}, pages 5562--5570, 2016.

\bibitem[Brox et~al.(2004)Brox, Bruhn, Papenberg, and Weickert]{brox2004high}
Thomas Brox, Andr{\'e}s Bruhn, Nils Papenberg, and Joachim Weickert.
\newblock High accuracy optical flow estimation based on a theory for warping.
\newblock In \emph{Proceedings of the European Conference on Computer Vision
  (ECCV)}, pages 25--36. Springer, 2004.

\bibitem[Carreira and Zisserman(2017)]{carreira2017quo}
Joao Carreira and Andrew Zisserman.
\newblock Quo vadis, action recognition? a new model and the kinetics dataset.
\newblock In \emph{Proceedings of the IEEE Conference on Computer Vision and
  Pattern Recognition (CVPR)}, pages 4724--4733. IEEE, 2017.

\bibitem[Castellano()]{sbd}
Brandon Castellano.
\newblock Pyscenedetect.
\newblock \url{https://github.com/Breakthrough/PySceneDetect/}.

\bibitem[Choi et~al.(2016)Choi, Oh, and So~Kweon]{choi2016video}
Jinsoo Choi, Tae-Hyun Oh, and In~So~Kweon.
\newblock Video-story composition via plot analysis.
\newblock In \emph{Proceedings of the IEEE Conference on Computer Vision and
  Pattern Recognition (CVPR)}, 2016.

\bibitem[Cohen et~al.(2003)Cohen, Sebe, Garg, Chen, and Huang]{cohen2003facial}
Ira Cohen, Nicu Sebe, Ashutosh Garg, Lawrence~S Chen, and Thomas~S Huang.
\newblock Facial expression recognition from video sequences: temporal and
  static modeling.
\newblock \emph{Computer Vision and image understanding}, 91\penalty0
  (1-2):\penalty0 160--187, 2003.

\bibitem[Essa and Pentland(1997)]{essa1997coding}
Irfan~A. Essa and Alex~Paul Pentland.
\newblock Coding, analysis, interpretation, and recognition of facial
  expressions.
\newblock \emph{IEEE transactions on pattern analysis and machine
  intelligence}, 19\penalty0 (7):\penalty0 757--763, 1997.

\bibitem[Fleet and Weiss(2006)]{fleet2006optical}
David Fleet and Yair Weiss.
\newblock Optical flow estimation.
\newblock In \emph{Handbook of mathematical models in computer vision}, pages
  237--257. Springer, 2006.

\bibitem[Freytag(1896)]{freytag1896freytag}
Gustav Freytag.
\newblock \emph{Freytag's technique of the drama: an exposition of dramatic
  composition and art}.
\newblock Scholarly Press, 1896.

\bibitem[Gkioxari and Malik(2015)]{gkioxari2015finding}
Georgia Gkioxari and Jitendra Malik.
\newblock Finding action tubes.
\newblock In \emph{Proceedings of the IEEE Conference on Computer Vision and
  Pattern Recognition (CVPR)}, pages 759--768. IEEE, 2015.

\bibitem[He et~al.(2016)He, Zhang, Ren, and Sun]{he2016deep}
Kaiming He, Xiangyu Zhang, Shaoqing Ren, and Jian Sun.
\newblock Deep residual learning for image recognition.
\newblock In \emph{Proceedings of the IEEE Conference on Computer Vision and
  Pattern Recognition (CVPR)}, pages 770--778, 2016.

\bibitem[Huang et~al.(2017)Huang, Rathod, Sun, Zhu, Korattikara, Fathi,
  Fischer, Wojna, Song, Guadarrama, and Murphy]{Huang_2017_CVPR}
Jonathan Huang, Vivek Rathod, Chen Sun, Menglong Zhu, Anoop Korattikara,
  Alireza Fathi, Ian Fischer, Zbigniew Wojna, Yang Song, Sergio Guadarrama, and
  Kevin Murphy.
\newblock Speed/accuracy trade-offs for modern convolutional object detectors.
\newblock In \emph{Proceedings of the IEEE Conference on Computer Vision and
  Pattern Recognition (CVPR)}, July 2017.

\bibitem[Hussain et~al.(2017)Hussain, Zhang, Zhang, Ye, Thomas, Agha, Ong, and
  Kovashka]{hussain2017automatic}
Zaeem Hussain, Mingda Zhang, Xiaozhong Zhang, Keren Ye, Christopher Thomas,
  Zuha Agha, Nathan Ong, and Adriana Kovashka.
\newblock Automatic understanding of image and video advertisements.
\newblock In \emph{Proceedings of the IEEE Conference on Computer Vision and
  Pattern Recognition (CVPR)}, pages 1100--1110. IEEE, 2017.

\bibitem[Ilg et~al.(2017)Ilg, Mayer, Saikia, Keuper, Dosovitskiy, and
  Brox]{ilg2017flownet}
Eddy Ilg, Nikolaus Mayer, Tonmoy Saikia, Margret Keuper, Alexey Dosovitskiy,
  and Thomas Brox.
\newblock Flownet 2.0: Evolution of optical flow estimation with deep networks.
\newblock In \emph{Proceedings of the IEEE Conference on Computer Vision and
  Pattern Recognition (CVPR)}, volume~2, 2017.

\bibitem[Joo et~al.(2014)Joo, Li, Steen, and Zhu]{joo2014visual}
Jungseock Joo, Weixin Li, Francis~F Steen, and Song-Chun Zhu.
\newblock Visual persuasion: Inferring communicative intents of images.
\newblock In \emph{Proceedings of the IEEE Conference on Computer Vision and
  Pattern Recognition (CVPR)}, pages 216--223, 2014.

\bibitem[Joo et~al.(2015)Joo, Steen, and Zhu]{joo2015automated}
Jungseock Joo, Francis~F Steen, and Song-Chun Zhu.
\newblock Automated facial trait judgment and election outcome prediction:
  Social dimensions of face.
\newblock In \emph{Proceedings of the IEEE International Conference on Computer
  Vision (ICCV)}, pages 3712--3720, 2015.

\bibitem[Kanade et~al.(2000)Kanade, Cohn, and Tian]{kanade2000comprehensive}
Takeo Kanade, Jeffrey~F Cohn, and Yingli Tian.
\newblock Comprehensive database for facial expression analysis.
\newblock In \emph{Automatic Face and Gesture Recognition, 2000. Proceedings.
  Fourth IEEE International Conference on}, pages 46--53. IEEE, 2000.

\bibitem[Kosti et~al.(2017)Kosti, Alvarez, Recasens, and
  Lapedriza]{kosti2017emotic}
Ronak Kosti, Jose~M Alvarez, Adria Recasens, and Agata Lapedriza.
\newblock Emotic: Emotions in context dataset.
\newblock In \emph{Computer Vision and Pattern Recognition Workshops (CVPRW),
  2017 IEEE Conference on}, pages 2309--2317. IEEE, 2017.

\bibitem[Lin et~al.(2014)Lin, Maire, Belongie, Hays, Perona, Ramanan,
  Doll{\'a}r, and Zitnick]{lin2014microsoft}
Tsung-Yi Lin, Michael Maire, Serge Belongie, James Hays, Pietro Perona, Deva
  Ramanan, Piotr Doll{\'a}r, and C~Lawrence Zitnick.
\newblock Microsoft coco: Common objects in context.
\newblock In \emph{Proceedings of the European Conference on Computer Vision
  (ECCV)}, pages 740--755. Springer, 2014.

\bibitem[Liu et~al.(2008)Liu, Li, Zhang, Tang, Song, and
  Yang]{liu2008innovative}
Anan Liu, Jintao Li, Yongdong Zhang, Sheng Tang, Yan Song, and Zhaoxuan Yang.
\newblock An innovative model of tempo and its application in action scene
  detection for movie analysis.
\newblock In \emph{WACV}, 2008.

\bibitem[Mahadevan et~al.(2010)Mahadevan, Li, Bhalodia, and
  Vasconcelos]{mahadevan2010anomaly}
Vijay Mahadevan, Weixin Li, Viral Bhalodia, and Nuno Vasconcelos.
\newblock Anomaly detection in crowded scenes.
\newblock In \emph{Proceedings of the IEEE Conference on Computer Vision and
  Pattern Recognition (CVPR)}, pages 1975--1981. IEEE, 2010.

\bibitem[Mayer et~al.(2016)Mayer, Ilg, Hausser, Fischer, Cremers, Dosovitskiy,
  and Brox]{mayer2016large}
Nikolaus Mayer, Eddy Ilg, Philip Hausser, Philipp Fischer, Daniel Cremers,
  Alexey Dosovitskiy, and Thomas Brox.
\newblock A large dataset to train convolutional networks for disparity,
  optical flow, and scene flow estimation.
\newblock In \emph{Proceedings of the IEEE Conference on Computer Vision and
  Pattern Recognition (CVPR)}, pages 4040--4048, 2016.

\bibitem[Mollahosseini et~al.(2017)Mollahosseini, Hasani, and
  Mahoor]{mollahosseini2017affectnet}
Ali Mollahosseini, Behzad Hasani, and Mohammad~H Mahoor.
\newblock Affectnet: A database for facial expression, valence, and arousal
  computing in the wild.
\newblock \emph{IEEE Transactions on Affective Computing}, 2017.

\bibitem[Na et~al.(2017)Na, Lee, Kim, and Kim]{Na_2017_ICCV}
Seil Na, Sangho Lee, Jisung Kim, and Gunhee Kim.
\newblock A read-write memory network for movie story understanding.
\newblock In \emph{Proceedings of the IEEE International Conference on Computer
  Vision (ICCV)}, Oct 2017.

\bibitem[Newell et~al.(2016)Newell, Yang, and Deng]{newell2016stacked}
Alejandro Newell, Kaiyu Yang, and Jia Deng.
\newblock Stacked hourglass networks for human pose estimation.
\newblock In \emph{Proceedings of the European Conference on Computer Vision
  (ECCV)}, pages 483--499. Springer, 2016.

\bibitem[Ranjan and Black(2017)]{ranjan2017optical}
Anurag Ranjan and Michael~J Black.
\newblock Optical flow estimation using a spatial pyramid network.
\newblock In \emph{Proceedings of the IEEE Conference on Computer Vision and
  Pattern Recognition (CVPR)}, volume~2, 2017.

\bibitem[Rasheed and Shah(2002)]{rasheed2002movie}
Zeeshan Rasheed and Mubarak Shah.
\newblock Movie genre classification by exploiting audio-visual features of
  previews.
\newblock In \emph{ICPR}, 2002.

\bibitem[Russakovsky et~al.(2015)Russakovsky, Deng, Su, Krause, Satheesh, Ma,
  Huang, Karpathy, Khosla, Bernstein, et~al.]{russakovsky2015imagenet}
Olga Russakovsky, Jia Deng, Hao Su, Jonathan Krause, Sanjeev Satheesh, Sean Ma,
  Zhiheng Huang, Andrej Karpathy, Aditya Khosla, Michael Bernstein, et~al.
\newblock Imagenet large scale visual recognition challenge.
\newblock \emph{International Journal of Computer Vision}, 115\penalty0
  (3):\penalty0 211--252, 2015.

\bibitem[Shotton et~al.(2013)Shotton, Girshick, Fitzgibbon, Sharp, Cook,
  Finocchio, Moore, Kohli, Criminisi, Kipman, et~al.]{shotton2013efficient}
Jamie Shotton, Ross Girshick, Andrew Fitzgibbon, Toby Sharp, Mat Cook, Mark
  Finocchio, Richard Moore, Pushmeet Kohli, Antonio Criminisi, Alex Kipman,
  et~al.
\newblock Efficient human pose estimation from single depth images.
\newblock \emph{IEEE Transactions on Pattern Analysis and Machine
  Intelligence}, 35\penalty0 (12):\penalty0 2821--2840, 2013.

\bibitem[Sun et~al.(2010)Sun, Roth, and Black]{sun2010secrets}
Deqing Sun, Stefan Roth, and Michael~J Black.
\newblock Secrets of optical flow estimation and their principles.
\newblock In \emph{Proceedings of the IEEE Conference on Computer Vision and
  Pattern Recognition (CVPR)}, pages 2432--2439. IEEE, 2010.

\bibitem[Szegedy et~al.(2016)Szegedy, Vanhoucke, Ioffe, Shlens, and
  Wojna]{szegedy2016rethinking}
Christian Szegedy, Vincent Vanhoucke, Sergey Ioffe, Jon Shlens, and Zbigniew
  Wojna.
\newblock Rethinking the inception architecture for computer vision.
\newblock In \emph{Proceedings of the IEEE Conference on Computer Vision and
  Pattern Recognition (CVPR)}, pages 2818--2826, 2016.

\bibitem[Tapaswi et~al.(2016)Tapaswi, Zhu, Stiefelhagen, Torralba, Urtasun, and
  Fidler]{tapaswi2016movieqa}
Makarand Tapaswi, Yukun Zhu, Rainer Stiefelhagen, Antonio Torralba, Raquel
  Urtasun, and Sanja Fidler.
\newblock Movieqa: Understanding stories in movies through question-answering.
\newblock In \emph{Proceedings of the IEEE Conference on Computer Vision and
  Pattern Recognition (CVPR)}, pages 4631--4640, 2016.

\bibitem[Teney et~al.(2017)Teney, Anderson, He, and Hengel]{teney2017tips}
Damien Teney, Peter Anderson, Xiaodong He, and Anton van~den Hengel.
\newblock Tips and tricks for visual question answering: Learnings from the
  2017 challenge.
\newblock \emph{arXiv preprint arXiv:1708.02711}, 2017.

\bibitem[Toshev and Szegedy(2014)]{toshev2014deeppose}
Alexander Toshev and Christian Szegedy.
\newblock Deeppose: Human pose estimation via deep neural networks.
\newblock In \emph{Proceedings of the IEEE Conference on Computer Vision and
  Pattern Recognition (CVPR)}, pages 1653--1660, 2014.

\bibitem[Tran et~al.(2015)Tran, Bourdev, Fergus, Torresani, and
  Paluri]{tran2015learning}
Du~Tran, Lubomir Bourdev, Rob Fergus, Lorenzo Torresani, and Manohar Paluri.
\newblock Learning spatiotemporal features with 3d convolutional networks.
\newblock In \emph{Proceedings of the IEEE International Conference on Computer
  Vision (ICCV)}, pages 4489--4497. IEEE, 2015.

\bibitem[Wang et~al.(2015)Wang, Qiao, and Tang]{wang2015action}
Limin Wang, Yu~Qiao, and Xiaoou Tang.
\newblock Action recognition with trajectory-pooled deep-convolutional
  descriptors.
\newblock In \emph{Proceedings of the IEEE Conference on Computer Vision and
  Pattern Recognition (CVPR)}, pages 4305--4314, 2015.

\bibitem[Wang et~al.(2017)Wang, Feng, Hong, Berger, and Luo]{wang2017polarized}
Yu~Wang, Yang Feng, Zhe Hong, Ryan Berger, and Jiebo Luo.
\newblock How polarized have we become? a multimodal classification of trump
  followers and clinton followers.
\newblock In \emph{International Conference on Social Informatics}, pages
  440--456. Springer, 2017.

\bibitem[Weng et~al.(2009)Weng, Chu, and Wu]{weng2009rolenet}
Chung-Yi Weng, Wei-Ta Chu, and Ja-Ling Wu.
\newblock Rolenet: Movie analysis from the perspective of social networks.
\newblock \emph{IEEE Transactions on Multimedia}, 11\penalty0 (2):\penalty0
  256--271, 2009.

\bibitem[Won et~al.(2017)Won, Steinert-Threlkeld, and Joo]{won2017protest}
Donghyeon Won, Zachary~C Steinert-Threlkeld, and Jungseock Joo.
\newblock Protest activity detection and perceived violence estimation from
  social media images.
\newblock In \emph{Proceedings of the 2017 ACM on Multimedia Conference}, pages
  786--794. ACM, 2017.

\bibitem[Ye and Kovashka(2018)]{ye2018advise}
Keren Ye and Adriana Kovashka.
\newblock Advise: Symbolism and external knowledge for decoding advertisements.
\newblock In \emph{European Conference on Computer Vision (ECCV)}. Springer,
  2018.

\bibitem[Yeung et~al.(2016)Yeung, Russakovsky, Mori, and Fei-Fei]{yeung2016end}
Serena Yeung, Olga Russakovsky, Greg Mori, and Li~Fei-Fei.
\newblock End-to-end learning of action detection from frame glimpses in
  videos.
\newblock In \emph{Proceedings of the IEEE Conference on Computer Vision and
  Pattern Recognition (CVPR)}, pages 2678--2687, 2016.

\bibitem[Young(2008)]{young2008advertising}
Charles~E Young.
\newblock \emph{The advertising research handbook}.
\newblock Ideas in Flight, 2008.

\bibitem[Zhou et~al.(2017)Zhou, Lapedriza, Khosla, Oliva, and
  Torralba]{zhou2017places}
Bolei Zhou, Agata Lapedriza, Aditya Khosla, Aude Oliva, and Antonio Torralba.
\newblock Places: A 10 million image database for scene recognition.
\newblock \emph{IEEE Transactions on Pattern Analysis and Machine
  Intelligence}, 2017.

\end{thebibliography}
\end{document}